\documentclass[journal]{IEEEtran}
\usepackage{cite}
\usepackage{graphicx}
\usepackage{booktabs}
\usepackage{subfigure}
\usepackage{algorithm}
\usepackage{algpseudocode}
\usepackage[colorlinks,linkcolor=red,citecolor=green,urlcolor=blue]{hyperref}
\usepackage{amsthm,amsmath,amssymb}
\usepackage{bm}
\usepackage{multirow}
\usepackage{bbding}
\usepackage{pifont}
\usepackage{wasysym}
\usepackage{tabularx}
\usepackage{makecell}

\makeatletter
\def\ps@IEEEtitlepagestyle{%
  \def\@oddfoot{\mycopyrightnotice}%
  \def\@evenfoot{}%
}
\def\mycopyrightnotice{%
  {\hfill \scriptsize {This work has been submitted to the IEEE for possible publication.
  Copyright may be transferred without notice, after which this version may no longer be accessible.}\hfill}

}
\makeatother

\ifCLASSINFOpdf
\else
\fi
\begin{document}

\title{Multi-modal Large Language Model Enhanced Pseudo 3D Perception Framework for Visual Commonsense Reasoning}

\author{Jian~Zhu,
        ~Hanli~Wang,~\IEEEmembership{Senior Member,~IEEE},
        ~and Miaojing Shi, ~\IEEEmembership{Senior Member,~IEEE}

        \thanks{
         \emph{Corresponding author: Hanli~Wang.}
         }

         \thanks{J.~Zhu and H.~Wang are with the Department of Computer Science \& Technology and the Key Laboratory of Embedded System \& Service Computing (Ministry of Education), Tongji University, Shanghai 200092, P. R. China (e-mail: jianzhu@tongji.edu.cn, hanliwang@tongji.edu.cn).

         Miaojing Shi is with the College of Electronic and Information Engineering, Tongji University, Shanghai 201804, P. R. China
         (e-mail: mshi@tongji.edu.cn).}
         }

\maketitle

\begin{abstract}
  The visual commonsense reasoning (VCR) task is to choose an answer and provide a justifying rationale based on the given image and textural question. Representative works first recognize objects in images and then associate them with key words in texts. However, existing approaches do not consider exact positions of objects in a human-like three-dimensional (3D) manner, making them incompetent to accurately distinguish objects and understand visual relation. Recently, multi-modal large language models (MLLMs) have been used as powerful tools for several multi-modal tasks but not for VCR yet, which requires elaborate reasoning on specific visual objects referred by texts. In light of the above, an MLLM enhanced pseudo 3D perception framework is designed for VCR. Specifically, we first demonstrate that the relation between objects is relevant to object depths in images, and hence introduce object depth into VCR frameworks to infer 3D positions of objects in images. Then, a depth-aware Transformer is proposed to encode depth differences between objects into the attention mechanism of Transformer to discriminatively associate objects with visual scenes guided by depth. To further associate the answer with the depth of visual scene, each word in the answer is tagged with a pseudo depth to realize depth-aware association between answer words and objects. On the other hand, BLIP-2 as an MLLM is employed to process images and texts, and the referring expressions in texts involving specific visual objects are modified with linguistic object labels to serve as comprehensible MLLM inputs. Finally, a parameter optimization technique is devised to fully consider the quality of data batches based on multi-level reasoning confidence. Experiments on the VCR dataset demonstrate the superiority of the proposed framework over state-of-the-art approaches.
\end{abstract}


%
\IEEEpeerreviewmaketitle

\section{Introduction}
\label{sec:intro}

In recent years, with the rapid growth of vision-and-language data, a number of challenging tasks have been studied such as referring expressions~\cite{Shang2023Cross, Zhu2022Multi}, image and video captioning~\cite{Xu2022Bridging, Yan2022Video}, visual question answering~(VQA)~\cite{Zhao2023Toward, Guo2021Re} and visual commonsense reasoning~(VCR)~\cite{Zellers2019From, Zhang2022Explicit}. These tasks require cross-modal intelligence to not only recognize entities in data but also understand their intrinsic interactions in varying degrees. Among them, VCR is a reasoning task aiming to choose an answer and provide a rationale justifying the answer based on the given image and question, which has many application potentials, \emph{e.g.}, robot dialogue system~\cite{Zhu2023MiniGPT} and private virtual assistant~\cite{Liu2023Visual}.

To tackle the VCR task, several approaches adopted multi-branch models to associate cross-modal data and fuse obtained features for later reasoning~\cite{Zellers2019From, Yu2019Heterogeneous, Zhang2022Explicit}. For instance, Zeller~\emph{et al.}~\cite{Zellers2019From} used long-short term memory~(LSTM) and attention mechanism to contextualize answer with image or question. Zhang~\emph{et al.}~\cite{Zhang2022Explicit} associated structured syntactic triplets of different sentences with visual graphs for reasoning. Yu~\emph{et al.}~\cite{Yu2019Heterogeneous} constructed heterogeneous graphs to model correlations between different domains. In the VCR task, the target image contains the facts that can be used to do reasoning. Therefore, it is crucial to extract discriminative visual features and properly associate them with languages. Lin~\emph{et al.}~\cite{Lin2019TAB} adopted an object detector integrating attributes such as colors, texture and size to enhance visual features for VCR, and this was also used in several follow-up works~\cite{Zhang2022Explicit, Zhang2021Multi}.

\begin{figure}[htbp]
   \centering
   \includegraphics[width=0.95\linewidth]{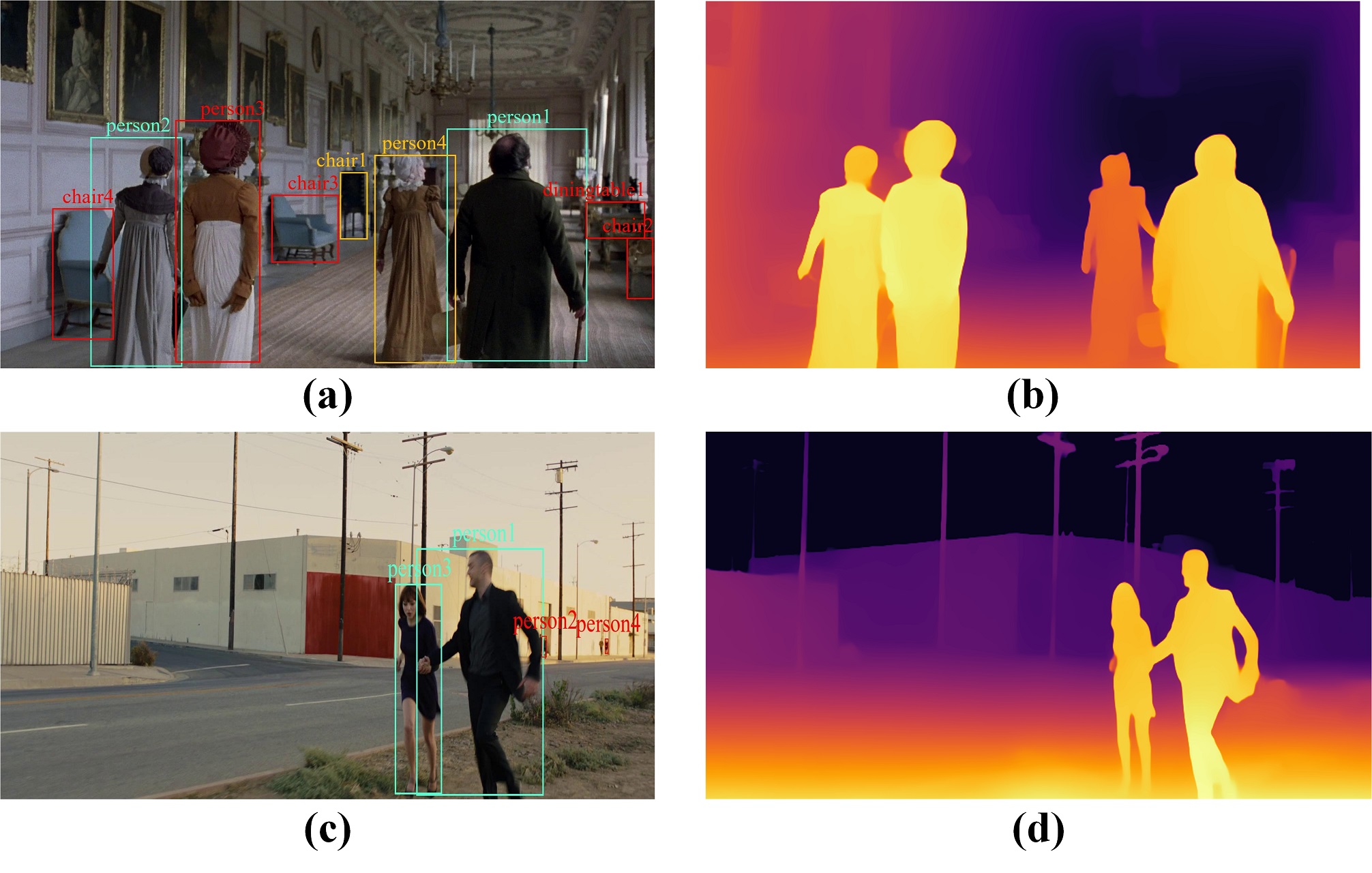}
   \caption{Example of RGB images (left) and the corresponding depth images (right). In the depth images, the pixel appears from yellow to black as the distance becomes farther. With the help of image depth, objects can be perceived in a human-like 3D manner.}
   \label{fig1}
\end{figure}

The aforementioned works focus on extracting a set of objects and associating them with words, whilst the positions of objects in the scene are ignored. However, positions are generally important to understand the relation between objects. Intuitively, the two-dimensional~(2D) bounding box produced by an object detector can be used to represent positions of objects in an image; nevertheless, we argue that this is insufficient to infer object relation. As shown in Fig.~\ref{fig1}(a), ``person 4'' is close to ``chair 1'' with respect to the 2D coordinates in the image while ``person 4'' is actually far away from ``chair 1'' according to human perception. An object requires typically a set of three-dimensional~(3D) coordinates to completely represent the position in the real world. Compared with the 2D coordinates in images, there is an additional dimension measuring the distance from the object to the image plane in the 3D coordinates, which can be converted to image depth~\cite{Ranftl2021Vision}. Recently, with the development of monocular depth estimation, image depth can be easily obtained from a monocular RGB image. Figure~\ref{fig1}(b) shows the depth image generated from Fig.~\ref{fig1}(a), which reflects that ``person 4'' is far away from ``chair 1'' with image depth considered.

Generally, semantic correlation can be used to represent the relation between objects. Besides semantic correlation, we believe that the relation between objects in images is also relevant to 3D distance and depth difference. On one hand, two objects with strong semantic correlation and close 3D distance usually have close relation, and this can be observed in Figs.~\ref{fig1}(c) and (d) that ``person 1'' holds the hand of ``person 3'' while ``person 1'' is irrelevant to ``person 2''. On the other hand, two objects with relatively far 3D distance but similar image depth may belong to the same visual scene. For instance, the 3D distance between ``person 1'' and ``person 2'' in Fig.~\ref{fig1}(a) is not close, but they belong to the same crowd and are with similar image depth values.

\begin{figure}[htbp]
   \centering
   \includegraphics[width=1.0\linewidth]{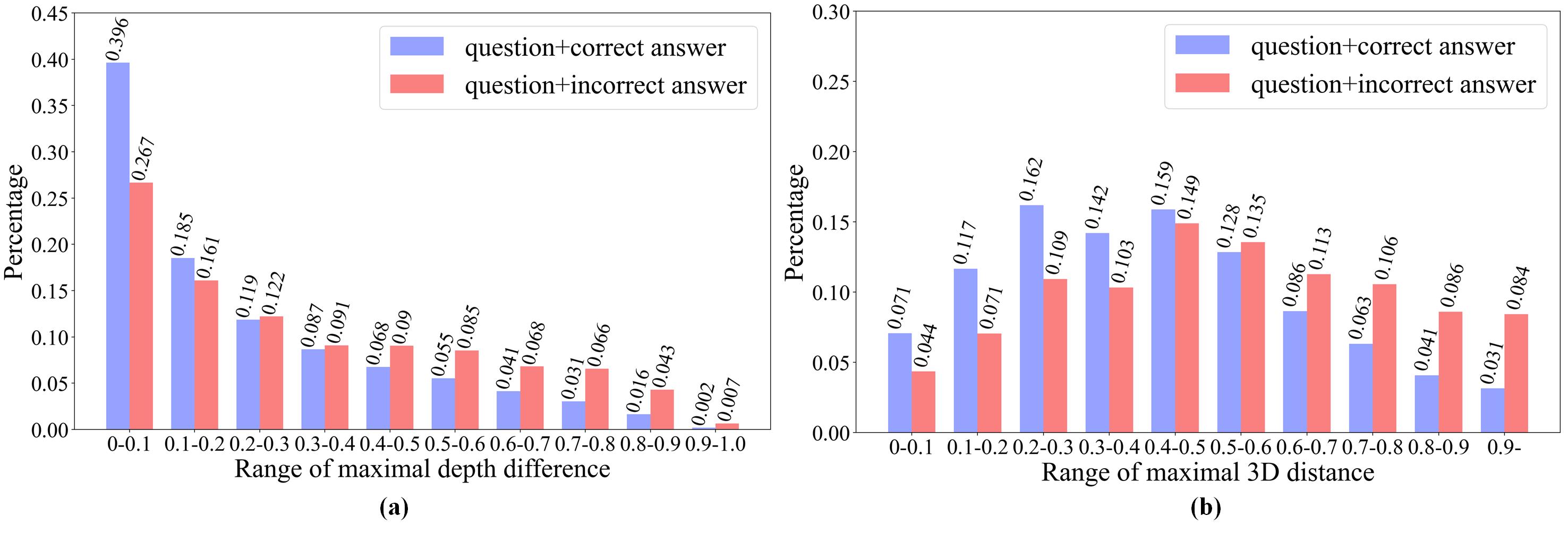}
   \caption{Statistics of the maximal depth difference and maximal 3D distance between all visual objects mentioned in the question and answer of VCR samples. Values of the depth, position $x$ and position $y$ in the image are normalized to the range $[0,1]$, so the value of 3D distance is in the range $[0, \sqrt{3}]$.}
   \label{fig2}
\end{figure}

To statistically analyze the correlation among image depth, 3D distance and object relation, the statistics of the maximal depth difference and maximal 3D distance between objects mentioned in the question and answer of VCR samples are shown in Fig.~\ref{fig2}. As for a VCR sample, the question and the correct answer can be combined as a text description depicting a visual scene happened in the image. The visual objects mentioned in the text description are key visual cues of the visual scene and thus viewed as highly related. As shown in Fig.~\ref{fig2}(a), the values of maximal depth difference between the related objects are in the small range from 0 to 0.3 among over 70\% VCR samples. This indicates that highly related objects are with similar depths. Motivated by this observation, we can associate more similar-depth objects not mentioned in the text description but relevant to a visual scene. By contrast, regarding the text descriptions using incorrect answers (\emph{i.e.}, the question plus the incorrect answer), some descriptions mention the related objects but picture the wrong visual scenes, and some others portray unrelated objects; the statistics shown in Fig.~\ref{fig2}(a) also verify that there are higher percentages of incorrect text descriptions with greater values of maximal depth difference as compared to the correct descriptions. Moreover, the statistics of the maximal 3D distance between the objects mentioned in text descriptions are shown in Fig.~\ref{fig2}(b), where the values of maximal 3D distance in 34.9\% correct descriptions are greater than 0.5 while 52.4\% for incorrect descriptions. This reveals that nearer objects measured in 3D distances are likely to be relevant, but it is insufficient to use 3D distance only to  judge object relation. It is reasonable to jointly consider both depth difference and 3D distance to infer object relation.

On the other hand, multi-modal large language models~(MLLMs) have been off-the-shelf tools to  align cross-modal features, which are trained on mass cross-modal data and usually used directly in many downstream tasks without extra training and fine-tuning. However, MLLMs cannot be directly applied to VCR as they are normally trained on the entire visual scene, while VCR requires elaborate reasoning on specific visual objects referred by texts. It is worth incorporating the MLLM's capability of aligning cross-modal features for the VCR task.

Inspired by the above motivations, an MLLM enhanced pseudo 3D perception framework, namely MEP3P, is designed for VCR in this work. The proposed MEP3P includes a pseudo 3D perception VCR framework~(P3PV) and an MLLM enhanced reasoning module~(MER). Similar to traditional VCR frameworks, the P3PV uses a two-branch~(answer-image branch and answer-question branch) Transformer~\cite{Vaswani2017Attention} architecture to independently associate the answer with the image or the question, and combines associated features of both branches for further reasoning. In the stage of visual representation, pseudo 3D positions of detected objects are calculated based on the depth image generated from the RGB image. The pseudo 3D positions and depth image features are used to enhance the original features extracted by a visual backbone. In the answer-image branch, depth-aware Transformer is designed to encode depth difference between objects to highlight object relation in the depicted visual scene. To further emphasize cross-modal relation and associate the answer with the visual scene guided by depth, each word in the answer is tagged with a pseudo depth to realize depth-aware association between answer words and objects. The proposed P3PV can independently do VCR, and the MER is further devised to provide better-aligned multi-modal features. The MER adopts the BLIP-2 model~\cite{Li2023BLIP} to process images and texts, and the referring expressions in texts involving specific visual objects are modified with linguistic object labels to serve as MLLM inputs. Moreover, the learning samples for VCR are generally diverse from each other, especially with additional depth information and cross-modal features. It is hard to use such diverse samples to train optimal VCR models. To address this issue, a parameter optimization technique is proposed to evaluate the quality of sample batches based on reasoning confidence and calculate optimization direction with sample quality considered.

The major contributions of this work are summarized below. First, we demonstrate that the relation between objects is relevant to object depth in images and introduce image depth to represent visual object in a pseudo-3D manner. Second, a depth-aware Transformer is proposed to encode depth difference into attention mechanism to associate objects and answer words with visual scenes guided by depth. Third, an MLLM enhanced reasoning module is designed to cooperate with the proposed pseudo 3D perception VCR framework by providing well-aligned cross-modal features and referring capability. Finally, the proposed MEP3P framework achieves competitive performances compared with several state-of-the-art methods on VCR benchmark datasets. The rest of this paper is organized as follows. We introduce the related work in Section~\ref{sect:rwork}. The proposed MEP3P framework is detailed in Section~\ref{sect:pm}. The experimental results are presented in Section~\ref{sect:exp}. Finally, we conclude this paper with a summary in Section~\ref{sect:cln}.

\begin{figure*}[htbp]
   \centering
   \includegraphics [width=0.95\linewidth] {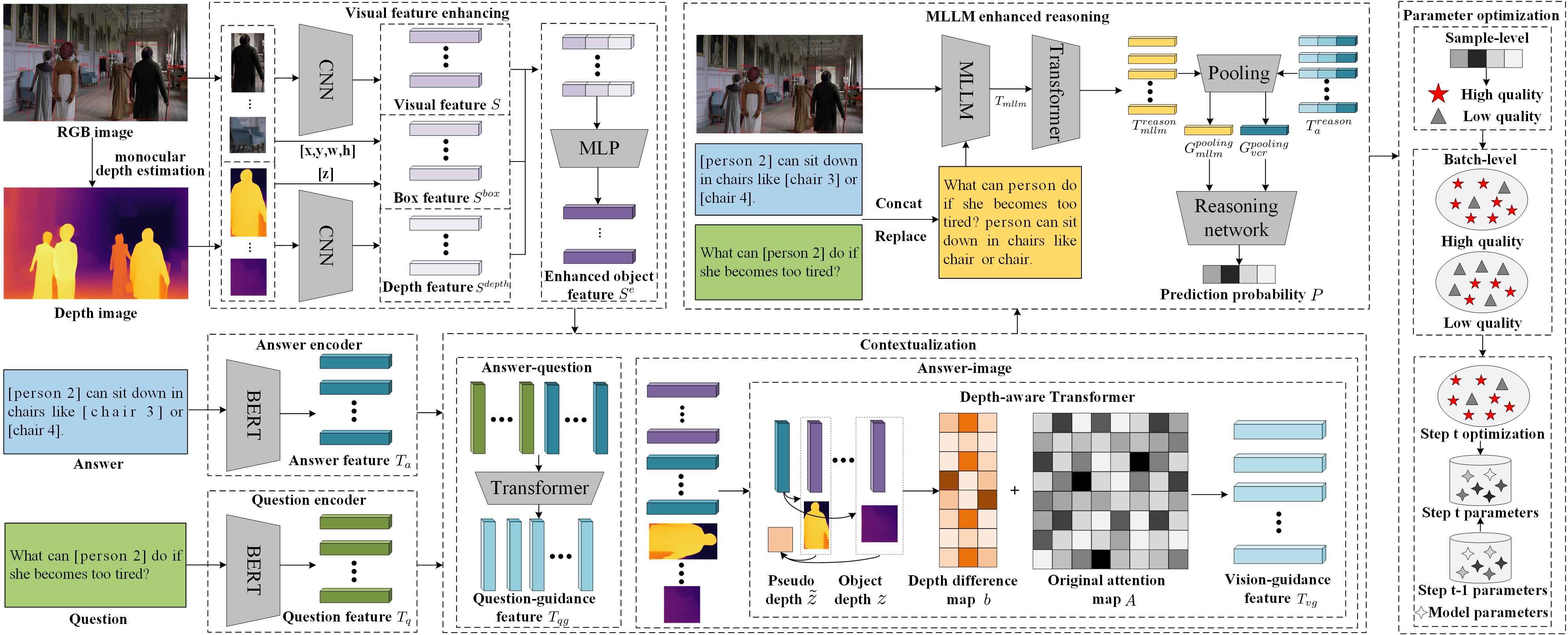}
   \caption{Illustration of the proposed MEP3P, which consists of four key parts: (1) visual feature enhancing, (2) vision-and-language contextualization via depth-aware Transformer, (3) MLLM enhanced reasoning, and (4) parameter optimization based on multi-level reasoning confidence. } \label{fig3}
\end{figure*}

\section{Related Work}
\label{sect:rwork}

\subsection{Visual Commonsense Reasoning}
\label{sect:rwork:vcr}

A main difficulty of VCR is how to sufficiently and properly represent visual cues from the image and associate them with linguistic data for reasoning. Previous methods can be roughly divided into two categories: sequential-data processing models~\cite{Zellers2019From, Lin2019TAB, Wen2020Multi, Su2020VL, Li2021UNIMO} and graph-based networks~\cite{Wu2019Connective, Yu2019Heterogeneous, Zhang2022Explicit, Zhu2023Multi}. In the early time, Zeller~\emph{et al.}~\cite{Zellers2019From} used LSTM with attention mechanism to contextualize sequential answer words with sequential question words or visual objects. Later, Lin~\emph{et al.}~\cite{Lin2019TAB} leveraged attributions of objects such as color and texture to enhance the visual features used in ~\cite{Zellers2019From}. Wen~\emph{et al.}~\cite{Wen2020Multi} adopted an additional linguistic knowledge base to transfer commonsense knowledge to LSTM for VCR. Su~\emph{et al.}~\cite{Su2020VL} used cross-modal sequential data consisting of visual objects and linguistic words from several cross-modal datasets to train a BERT-based model, then fine-tuned the model on VCR dataset. Li~\emph{et al.}~\cite{Li2021UNIMO} proposed a cross-modal contrastive learning method to achieve alignment of cross-modal data. Regarding graph representation and learning, Zhang~\emph{et al.}~\cite{Zhang2022Explicit} constructed visual and linguistic graphs respectively, and then fused them for cross-modal reasoning. Yu~\emph{et al.}~\cite{Yu2019Heterogeneous} built answer-vision and answer-question heterogeneous graphs to perform reasoning. Wu~\emph{et al.}~\cite{Wu2019Connective} developed visual neuron connectivity to model correlations of visual content, which was further improved by a connective cognition network with linguistic contextualization. Zhu~\emph{et al.}~\cite{Zhu2023Multi} built heterogeneous graphs with image region distance and linguistic word order considered, then learned graph evolution via graph Transformer. These methods focused on associating visual objects based on semantic features, but ignored their relative 3D position relation which is also important to facilitate the understanding of visual scene. In this work, image depth is introduced into VCR to represent the relation of visual objects in a pseudo-3D manner.

\subsection{Depth Estimation}
\label{sect:rwork:de}

Depth estimation from images is an important task in computer vision, which is widely used in simultaneous localization and mapping~(SLAM)~\cite{Zhang2022MOTSLAM}, object detection~\cite{Cao2022CMAN} and semantic segmentation~\cite{E2022Depth}. The approaches of depth estimation can be divided into binocular depth estimation approaches such as~\cite{Pilzer2020Progressive} and monocular depth estimation approaches such as~\cite{Ranftl2021Vision}. Compared with the binocular camera system, the monocular camera system is more convenient and less expensive to be deployed. Zhang~\emph{et al.}~\cite{Zhang2022MOTSLAM} presented a dynamic visual SLAM system tracking both poses and bounding boxes of dynamic object, where dynamic features were fetched by monocular depth estimation. Cao~\emph{et al.}~\cite{Cao2022CMAN} encoded structure correlations from monocular depth data, and embedded them with appearance information learnt from RGB data. Ergl~\emph{et al.}~\cite{E2022Depth} extracted the 3D geometric structure of visual scene based on monocular depth map, which guided segmentation model to learn object boundary. These works mainly used depth maps to assist visual tasks, while few efforts were devoted into more complicated cross-modal tasks. In this work, image depth is investigated in the cross-modal VCR task, where visual data interacts with linguistic data to realize visual understanding.

\subsection{Multi-modal Large Language Model}
\label{sect:rwork:mllm}

The typical functions of MLLMs such as~\cite{Li2023BLIP, Zhu2023MiniGPT, Liu2023Visual} are to do vision-and-language alignment and decode aligned representations for large language model~(LLM). MLLMs have been trained on mass cross-modal data, and can be directly used in many down-stream tasks such as visual question answering, image captioning and visual dialogue without extra training and fine-tuning. However, VCR is a more complicated task with referring expressions, and MLLM cannot directly tackle it. To make use of the strength of MLLM, we utilize the BLIP-2 model~\cite{Li2023BLIP} to extract well-aligned cross-modal features from image and text. Then, the aligned features are adopted to assist the P3PV to understand visual scene.

\section{Proposed MLLM Enhanced Pseudo 3D Perception Framework}
\label{sect:pm}

\subsection{Overview}
\label{sect:pp:overview}

The VCR task is composed of the following three subtasks:

1) $Q\rightarrow A$: Given an image and a question, a model chooses the right option from four answers.

2) $QA\rightarrow R$: Given an image, a question and the correct answer, a model chooses the right option from four rationales. In this subtask, the original question and the correct answer are concatenated as the input question of the model, and the rationales serve as the input answers of the model.

3) $Q\rightarrow AR$: Given an image and a question, a model chooses both the right answer and rationale.

The $Q\rightarrow AR$ task can be decomposed into $Q\rightarrow A$ plus $QA\rightarrow R$, and the result of $Q\rightarrow AR$ is achieved via combining the results of $Q\rightarrow A$ and $QA\rightarrow R$.

The overall architecture of the proposed framework MEP3P is illustrated in Fig.~\ref{fig3}. The features of image regions, question and answer are firstly extracted by pre-trained networks. Then, the original visual features are enhanced by image depth features along with pseudo 3D positions. The MEP3P utilizes plain Transformer and depth-aware Transformer to do contextualization in answer-question and answer-image branches, respectively.
The aligned semantic representations of the overall visual scene conditioned on the text are further obtained via the BLIP-2 model~\cite{Li2023BLIP} for reasoning. Finally, model parameters are optimized in consideration of the quality of sample batches evaluated with multi-level reasoning confidence.

\subsection{Depth-enhanced Visual Feature}
\label{sect:pp:vfeid}

For a question with $N_{q}$ words $U_{q}=\{u_{i}^{q}\}_{i=1}^{N_{q}}$ or an answer with $N_{a}$ words $U_{a}=\{u_{i}^{a}\}_{i=1}^{N_{a}}$, a pre-trained BERT and a subsequent bidirectional LSTM are used to extract the linguistic features, denoted as $T_{q}=\{t_{i}^{q}\}_{i=1}^{N_{q}}, t_{i}^{q} \in \mathbb{R}^{d}$ for the question and $T_{a}=\{t_{i}^{a}\}_{i=1}^{N_{a}}, t_{i}^{a} \in \mathbb{R}^{d}$ for the answer~($ d $ is the feature dimension). Visual features are enhanced by pseudo 3D positions and depth image features. With regard to $N_{o}$ objects $O=\{o_{i}\}_{i=1}^{N_{o}}$ in a RGB image $I$, a pre-trained CNN is utilized to extract the original visual features $S=\{s_{i}\}_{i=1}^{N_{o}}, s_{i} \in \mathbb{R}^{d}$. The depth image $I_{D}$ is generated via the monocular depth estimation model~\cite{Ranftl2021Vision} applied on the original RGB image $I$. Each object $o_{i}$ corresponds to a bounding box denoted as $(x_{i}, y_{i}, w_{i}, h_{i})$, where $(x_{i},y_{i})$ represents the 2D coordinate of the box center, $ w_{i} $ and $ h_{i} $ represent the width and height of the box. The depth image is aligned at pixel positions with the RGB image, and the bounding box can also be used to indicate the 2D position of the object in the depth image. The object is typically in the center of the bounding box with a few irrelevant pixels on the edge, and the change of depth within object is not significant. Therefore, the pixel value $z_{i} $ at the position of box center $ (x_{i}, y_{i}) $ in the depth image can be viewed as the depth coordinate of the object. Then, the pseudo 3D position feature is characterized by the box feature $ s_{i}^{box} $ consisting of five elements as
\begin{equation}
s_{i}^{box}= [x_{i}, y_{i}, z_{i}, w_{i}, h_{i}],
\end{equation}
where $x_{i}, y_{i}, z_{i}, w_{i}, h_{i}$ are all normalized in the range $[0,1]$.

To integrate depth information more accurately, a typical CNN is applied on the box region of the depth image to obtain the depth feature $s_{i}^{depth} \in \mathbb{R}^{d}$ of the object. The enhanced object feature $s_{i}^{e} \in \mathbb{R}^{d}$ is jointly learnt from $s_{i}$, $s_{i}^{box}$ and $s_{i}^{depth}$, which can be formulated as
\begin{equation}
s_{i}^{e} = fc(\mbox{concat}([s_{i}, s_{i}^{box}, s_{i}^{depth}])),
\label{eq2}
\end{equation}
where $fc(\cdot)$ is a fully-connected layer and concat($\cdot$) is the concatenating operator. The enhanced object feature with exact spatial information can represent the object more sufficiently, and provide spatial positional encoding for cross-modal Transformer in analogy with sequential positional encoding in plain Transformer.

\subsection{Depth-aware Transformer}
\label{sect:pf:DT}

After obtaining visual and linguistic features, we adopt Transformer-based models to associate them with each other. The plain Transformer~\cite{Vaswani2017Attention} is used in the answer-question branch while the depth-aware Transformer is proposed for the answer-image branch. Given a sequence $ X=\{x_{i}\}_{i=1}^{n} $ for plain Transformer, the attention mechanism can be formulated as
\begin{equation}
A = \frac{Q\cdot(K)^{T}}{\sqrt{d}},
\label{eq3}
\end{equation}
where $A\in \mathbb{R}^{n\times n}$ represents the attention weight matrix, $Q_{i}$ and $K_{i}$ are projections of $x_{i}$ using independent fully-connected layers. In the answer-image branch, the sequential input into depth-aware Transformer is composed of objects and answer words as $\{s_{1}^{e},\cdots,s_{N_{o}}^{e},t_{1}^{a},\cdots,t_{N_{a}}^{a}\}$.

\subsubsection{Depth-aware Attention between Objects}

The plain attention mechanism works mainly based on semantic similarities. However, as illustrated in Fig.~\ref{fig1}, the objects with similar visual appearances and quite different depth values may have discriminative semantic relation with another object. To better analyze the relation between objects in images, semantic correlation, 3D distance and depth difference can be jointly considered. On one hand, two objects with strong semantic correlation and close 3D distance usually have close relation. On the other hand, two objects with relatively far 3D distance but similar depth may belong to the same visual scene and have close relation. As for the objects $i$ and $j$, the enhanced object features $s_{i}^{e}$ and $s_{j}^{e}$ calculated by Eq.~\eqref{eq2} have considered the semantic information and 3D position. $Q$ and $K$ in the original attention mechanism as shown in Eq.~\eqref{eq3} are projected from the enhanced object features, so the attention weight $A_{ij}$ calculated by Eq.~\eqref{eq3} implicitly considers semantic similarity and 3D distance between the objects $i$ and $j$. To further consider the depth difference, we propose the following depth-aware attention as
\begin{equation}
A_{ij}^{depth} = A_{ij} + b_{ij},
\label{eq4}
\end{equation}
where $ b_{ij} $ is the term calculated based on depth difference to adjust $A_{ij}$. As for the objects $i$ and $j$, the depth difference is calculated by
\begin{equation}
\Delta z_{ij}=|z_{i}-z_{j}|,\; \Delta z_{ij} \in [0,1].
\end{equation}
Then, the adjustment term $ b_{ij} $ is defined as
\begin{equation}
b_{ij} = |A_{ij}|\cdot(e^{-\Delta z_{ij}}-\alpha),
\label{eq6}
\end{equation}
where $ \alpha $ is the depth bias. In this way, $ b_{ij} $ considers the scale of $ |A_{ij}| $ and grows from negative to positive as $ \Delta z_{ij} $ declines. As a result, the proposed depth-aware attention pays more attention to the two objects with similar depth values since they are more likely to belong to the same visual scene.

\subsubsection{Depth-aware Attention from Object to Word}

The answer is a text description depicting a specific visual scene and is thus related to the objects in the scene. As is demonstrated that the objects belonging to the same visual scene are likely to have similar depths, depth can be used to emphasize cross-modal relation from objects to the related words. Specifically, the word in the answer is viewed as implicitly corresponding to a depth level determined by visual scene, and a pseudo depth value is assigned to the answer word based on the related objects in the depth-aware Transformer. The pseudo depth $ \tilde{z_{i}} $ of the word $ i $ is calculated as
\begin{equation}
\tilde{z_{i}} = \sum_{0\leq m<N_{o}} \mbox{softmax}(A_{im})\cdot z_{m},
\end{equation}
where $z_{m}$ represents the depth of the $ m^{th} $ object in the input sequence, and $A_{ij}$ is the attention score from the object $m$ to the word $i$ obtained by the former Transformer layer. As a result, the depth difference $\Delta z_{ij}$ between word $i$ and object $j$ is obtained by
\begin{equation}
\Delta z_{ij}=|\tilde{z_{i}}-z_{j}|,
\end{equation}
which is used to calculate the adjustment term in Eq.~\eqref{eq6}, so the attention from object to word is also in a depth-aware manner similar to Eq.~\eqref{eq4}.

In the answer-image branch, the answer words and the visual objects are contextualized by several depth-aware Transformer layers, with the attention weight $A_{ij}^{depth}$ jointly considering semantic similarity and depth difference. The vision-guidance feature $T_{vg} \in \mathbb{R}^{N_{a}\times d}$ output by the final depth-aware Transformer layer is formulated as
\begin{gather}
T_{ai}=FFN(\mbox{softmax}(A^{depth})\cdot V),\\
T_{vg}=[T_{ai}^{N_{o}+1},\cdots,T_{ai}^{N_{o}+N_{a}}],
\end{gather}
where $V$ is the value matrix projected from the sequential input of depth-aware Transformer, and $FNN$ represents the feed forward network. In the answer-question branch, the plain Transformer~\cite{Vaswani2017Attention} is used to associate answer words with question words and the question-guidance feature $T_{qg} \in \mathbb{R}^{N_{a}\times d}$ is generated.

\subsection{MLLM Enhanced Reasoning}
\label{sect:pp:MER}

After the contextualization stage, the vision-guidance and the question-guidance features $T_{vg}$ and $T_{qg}$ are used for further reasoning, which are concatenated with the original answer word representation $T_{a}$ to obtain the sequential reasoning feature $T_{a}^{reason} \in \mathbb{R}^{N_{a} \times 3d}$ as
\begin{equation}
T_{a}^{reason} = \mbox{concat}([T_{a}, T_{vg}, T_{qg}]).
\end{equation}
The sequential reasoning feature is required to be pooled as a vector denoted as $G_{vcr}^{pooling}$ for the classification network, which is a two-layer fully-connected network adopted in~\cite{Zellers2019From, Lin2019TAB, Zhang2022Explicit} as well. In order to utilize the cross-modal alignment ability of MLLM, the BLIP-2~\cite{Li2023BLIP} model is employed to extract visual-and-linguistic features, denoted as $T_{mllm} \in \mathbb{R}^{N_{Qf}\times d}$, where $N_{Qf}$ is the number of tokens initialized in the Q-former module used in BLIP-2. However, the original question and answer of VCR contain referring expressions such as ``[person 1]'' in text, which cannot be directly processed by BLIP-2. To address this issue, the referring expression in text is replaced with semantic label. For instance, ``[person 1]'' and ``[person 2]'' are both replaced with the word ``person''. In the proposed MLLM enhanced reasoning module, the textural input to BLIP-2 is the combination of the question and answer pre-processed for the task $ Q\rightarrow A$, which includes additional rationale for the task $ QA\rightarrow R$. The output $T_{mllm}$ is projected by a Transformer layer as $T_{mllm}^{reason}$, then pooled as $G_{mllm}^{pooling}$. As a result, only the Transformer-based projection layer of MLLM needs to be trained. Finally, $G_{vcr}^{pooling}$ and $G_{mllm}^{pooling}$ are jointly fed into the classification network, then the output is activated by the softmax function to obtain the prediction probability of the four choices $P = [p_{1}, p_{2}, p_{3}, p_{4}]$ for answers (\emph{i.e.}, $Q\rightarrow A$) or rationales (\emph{i.e.}, $QA\rightarrow R$).

\subsection{Parameter Optimization Based on Reasoning Confidence}
\label{sect:pp:PERC}

As mentioned before, the diversity of VCR learning samples makes it difficult to train optimal VCR models. To tackle this problem, we propose a parameter optimization technique based on multi-level reasoning confidence, which considers the quality of sample batches evaluated by multi-level confidence to integrate multi-step optimization results.

To evaluate batch quality, sample quality should be firstly considered. Given a batch of $ N_{b} $ samples, the prediction on the batch is denoted as ${\{P_{i}\}}_{i=1}^{N_{b}}$. We define the following two kinds of high-quality: (1) sample-level high-quality means that the sample is predicted correctly with high reasoning confidence, conditioned on $\max(P_{i})>c_{sample}$, where $c_{sample}$ is the threshold to determine high reasoning confidence; and (2) batch-level high-quality means that the proportion of high-quality samples in the batch is high and the prediction accuracy on this batch $ batch_{cur} $ is higher than that on the previous batch $ batch_{pre}$, conditioned on $N_{sample\_h}/N_{b}>c_{batch}$ and $ acc(batch_{cur}) > acc(batch_{pre})$, where $ N_{sample\_h} $ is the number of high-quality samples in the batch, $c_{batch}$ is the threshold to judge high-quality batch and $acc(\cdot)$ stands for the prediction accuracy on the batch.

When optimizing the model parameters at the training step $t$, the set of parameter values denoted as $K_{t}$ is multiplied by a weight $ \beta $ according to batch quality, where $ \beta $ equals 1 for high-quality batches and $ N_{sample\_h}/N_{b} $ for low-quality batches. To further avoid unsatisfactory optimization caused by a specific batch, the finally optimized set of parameter values $K_{t}^{op}$ is obtained by a weighted summation of $K_{t}$ and the final set $K_{t-1}^{op}$ at the previous step $t-1$, as formulated as
\begin{equation}
K_{t}^{op} = \beta \cdot (1-\gamma)\cdot K_{t} + (1-\beta\cdot(1-\gamma))\cdot K_{t-1}^{op},
\label{eq12}
\end{equation}
where $\gamma$ is the integration weight parameter.

\subsection{Implementation}
\label{sect:pp:impl}

The proposed MEP3P is implemented on PyTorch. The models for the subtasks $Q\rightarrow A$ and $QA\rightarrow R$ share the identical architecture, which are trained separately. The prediction for the subtask $Q\rightarrow AR$ is the combination of the predictions for $Q\rightarrow A$ and $QA\rightarrow R$. For feature extraction of images, each object in the given image is represented as a 512-dimensional vector by the backbone~\cite{Lin2019TAB} based on ResNet101~\cite{He16Deep}. With regard to texts, each word is embedded as a 768-dimensional vector by the pre-trained BERT~\cite{devlin2019bert}, and the sequential linguistic features are fed into a single-layer bidirectional LSTM~(Bi-LSTM) model to generate a 512-dimensional vector for each word. The CNN network applied on depth image has three convolutional layers and a fully-connected layer to generate 512-dimensional depth features. The dimension of enhanced object feature is also 512. The Dropout rate is set to 0.3 in Bi-LSTM and 0.1 in Transformer. The Transformer-based module is trained by Noam~\cite{Vaswani2017Attention} while the other modules are trained by Adam~\cite{kingma2014adam} with the learning rate set to 0.0002. The depth bias parameter $\alpha$ in Eq.~\eqref{eq4} is set to 0.6, the integration weight parameter $\gamma$ in Eq.~\eqref{eq12} is set to 0.9998, the thresholds $c_{sample}$ and $c_{batch}$ are set to 0.55 and 0.6, respectively, the batch size is set to 96, and the model is trained for 30 epochs with early stopping.

\section{Experiment}
\label{sect:exp}

\subsection{Dataset}
\label{sect:exp:data}

Extensive experiments are conducted on the VCR benchmark dataset~\cite{Zellers2019From} to evaluate the proposed MEP3P. The dataset has 290k four-way multi-choice QA problems derived from 110k movie scenes. Compared with VQA datasets where answer is usually a single word, the answer and rationale in the VCR dataset are mixtures of visual and linguistic words. The average lengths of the answers and rationales are more than 7.5 words and 16 words, respectively. Following the data partition practice in~\cite{Zellers2019From}, the training set is composed of 80,418 images with 212,923 questions, and the validation set contains 9,929 images with 26,534 questions.

\subsection{Evaluation Metric and Baseline}
\label{sect:exp:Eval}

The VCR task adopts classification accuracy as the evaluation metric, which is a ratio of correctly classified samples to all test samples. The proposed MEP3P is compared with four kinds of methods: (1) text-only baselines, including BERT~\cite{devlin2019bert}, BERT (response only)~\cite{devlin2019bert}, ESIM+ELMO~\cite{chen2017enhanced} and LSTM+ELMO~\cite{chen2017enhanced}, which just utilize linguistic information and can be used to evaluate the importance of visual context in VCR; (2) VQA baselines, including RevisitedVQA~\cite{jabri2016revisiting}, BottomUpTopDown~\cite{Anderson2018Bottom}, MLB~\cite{kim2016hadamard} and MUTAN~\cite{ben2017mutan}, which are originally designed for VQA to process simple responses and modified to perform VCR; (3) VCR methods, including R2C~\cite{Zellers2019From}, CKRM~\cite{Wen2020Multi}, TAB-VCR~\cite{Lin2019TAB}, CCN~\cite{Wu2019Connective}, HGL~\cite{Yu2019Heterogeneous}, ECMR~\cite{Zhang2022Explicit}, MCC~\cite{Zhang2021Multi}, JAE~\cite{Li2022Joint}, VC R-CNN~\cite{Wang2020Visual} and CL-VCR~\cite{ye2021case}; (4) vision-and-language models, including VisualBERT~\cite{Li2019Visual}, VL-BERT~\cite{Su2020VL} and UNITER~\cite{chen2020UNITER}.

Below we briefly introduce the representative VCR and vision-and-language methods. R2C~\cite{Zellers2019From} uses LSTM with attention to contextualize sequential answer words with sequential question words or visual objects. CKRM~\cite{Wen2020Multi} is an attention-based model to transfer external knowledge into the VCR task. TAB-VCR~\cite{Lin2019TAB} adopts an object detector integrating attributes such as colors, texture and size to enhance visual features for VCR. CCN~\cite{Wu2019Connective} employs a connective cognition network and reorganizes visual neuron connectivity to do VCR. HGL~\cite{Yu2019Heterogeneous} constructs heterogeneous graphs to model correlations between image and text. ECMR~\cite{Zhang2022Explicit} associates structured syntactic triplets of different sentences with visual graphs for reasoning. MCC~\cite{Zhang2021Multi} generates counterfactual samples and employs contrastive learning to train VCR models. JAE~\cite{Li2022Joint} presents a plug-and-play knowledge distillation enhanced framework to perform VCR. VC R-CNN~\cite{Wang2020Visual} employs region-based CNN to perform causal intervention for visual feature enhancement. CL-VCR~\cite{ye2021case} adopts a curriculum-based masking approach to training model for VCR. VisualBERT~\cite{Li2019Visual} consists of a stack of Transformer layers that implicitly align the features of visual regions and linguistic words in image-text pairs with self-attention. VL-BERT~\cite{Su2020VL} adopts Transformer as the backbone to process visual and linguistic embedding features on several datasets to learn vision-and-language alignment. UNITER~\cite{chen2020UNITER} designs four pre-training tasks to do vision-and-language alignment.

\begin{table}[htbp]
\renewcommand\arraystretch{1.1}
\centering
\caption{Accuracy comparison for the three VCR subtasks achieved by competing methods on the VCR validation set.}
\newsavebox{\tableboxTabOV}
\begin{lrbox}{\tableboxTabOV}
\begin{tabular}{c|c|c|c}
\toprule[1pt]
Methods              & $Q\rightarrow A$ & $QA\rightarrow R$ & $Q\rightarrow AR$ \\
\hline
BERT~\cite{devlin2019bert}  & 53.8              & 64.1                      & 34.8            \\
BERT (response only)~\cite{devlin2019bert} & 27.6              & 26.3                     & 7.6            \\
ESIM+ELMO~\cite{chen2017enhanced}            & 45.8              & 55.0                      & 25.3            \\
LSTM+ELMO~\cite{chen2017enhanced}   & 28.1              & 28.7                     & 8.3            \\
\hline
RevisitedVQA~\cite{jabri2016revisiting}   & 39.4              & 34.0                      & 13.5            \\
BottomUpTopDown~\cite{Anderson2018Bottom}   & 42.8              & 25.1                      & 10.7           \\
MLB~\cite{kim2016hadamard}   & 45.5              & 36.1                      & 17.0           \\
MUTAN~\cite{ben2017mutan}   & 44.4              & 32.0                      & 14.6            \\
\hline
R2C~\cite{Zellers2019From}   & 63.8              & 67.2                     & 43.1            \\
CKRM~~\cite{Wen2020Multi}   & 66.2              & 68.5                      & 45.6           \\
TAB-VCR~\cite{Lin2019TAB}   & 69.9              & 72.2                      & 50.6            \\
CCN~\cite{Wu2019Connective}   & 67.4              & 70.6                      & 47.7            \\
HGL~\cite{Yu2019Heterogeneous}   & 69.4              & 70.6                      & 49.1            \\
ECMR~\cite{Zhang2022Explicit}   & 70.7              & 72.0                      & 51.1            \\
MCC~\cite{Zhang2021Multi}   & 71.7              & 73.4                      & 52.9            \\
JAE~\cite{Li2022Joint}   & 70.5              & 73.1                      & 51.8            \\
VC R-CNN~\cite{Wang2020Visual}   & 67.4              & 69.5                      & -            \\
CL-VCR~\cite{ye2021case}   & 69.9              & 70.6                      & -            \\
\hline
VisualBERT~\cite{Li2019Visual}    & 70.8              & 73.2                      & 52.2            \\
VL-BERT~\cite{Su2020VL}    & 73.8              &  74.4                     & 55.2            \\
UNITER~\cite{chen2020UNITER}    & 74.6              & {\bf 77.0}                      & 57.8            \\
\hline
MEP3P w/o MLLM         & 72.2            & 73.8                     & 53.5           \\
MEP3P           & {\bf 77.6}             & {\bf 77.0}                     & {\bf 60.0}           \\
\toprule[1pt]
\end{tabular}
\label{tab1}
\end{lrbox}
\scalebox{1.0}{\usebox{\tableboxTabOV}}
\end{table}

\subsection{Quantitative Result}
\label{sect:exp:QR}

The quantitative results achieved by the proposed MEP3P and several competing methods for the three sub-tasks in VCR are given in Table~\ref{tab1}. Regarding the text-only baselines, the performances are unsatisfactory without considering visual information. The VQA baselines additionally use visual information, but still cannot achieve satisfactory results since the textural expressions in VCR are more complicated than that of VQA. For the VCR methods, TAB-VCR~\cite{Lin2019TAB}, ECMR~\cite{Zhang2022Explicit} and MCC~\cite{Zhang2021Multi} use the visual features integrating attributes and achieve better performances than the other baselines such as R2C~\cite{Zellers2019From}, CKRM~\cite{Wen2020Multi}, CCN~\cite{Wu2019Connective} and HGL~\cite{Yu2019Heterogeneous} that employ plain visual features.

The proposed MEP3P w/o MLLM~(MEP3P without MLLM) model further introduces pseudo 3D position and image depth to enhance visual features, and obtains better performance with 72.2\% for the $Q\rightarrow A$ task, 73.8\% for the $QA\rightarrow R$ task, and 53.5\% for the $Q \rightarrow AR$ task, respectively. In comparison to MCC~\cite{Zhang2021Multi} which is trained via counterfactual samples and contrastive learning, the proposed MEP3P w/o MLLM adopts POMRC to optimize the framework, gaining the improvement of 0.6\% for the $Q \rightarrow AR$ task over MCC. As for the VC R-CNN~\cite{Wang2020Visual} method using causal intervention, the CL-VCR~\cite{ye2021case} method with robust training and JAE~\cite{Li2022Joint} adopting knowledge distillation, the proposed MEP3P w/o MLLM has superiority as well.
With regard to the vision-and-language models typically requiring 16 Tesla V100 GPUs and several hundred hours to train for obtaining aligned features, the proposed MEP3P method only needs 2 Tesla V100 GPUs to train for about 30 hours with MLLM parameters frozen. With the assistance of MLLM, the proposed MEP3P can also use well-aligned cross-modal features, and gain the improvement of 2.2\% for the $Q \rightarrow AR$ task over UNITER.

\subsection{Ablation Study}
\label{sect:exp:abStu}

To evaluate the effectiveness of the proposed modules of visual feature enhancing~(VFE), depth-aware Transformer~(DT), parameter optimization based on multi-level reasoning confidence~(POMRC) and MLLM enhanced reasoning~(MER), several models are designed to perform ablation study, including (1) {\bf Base}: a variant of the R2C model~\cite{Zellers2019From} replaces the backbone with ResNet101 and uses plain Transformer to do contextualization and reasoning instead of LSTM; (2) {\bf Base+VFE2D}: a variant of the base model uses the visual features enhanced by 2D positions obtained in the RGB image, which is to evaluate the effect of 2D object positions; (3) {\bf Base+VFE}: a variant model of MEP3P w/o MLLM adopts the visual features enhanced by pseudo 3D positions and depth image features, which is to evaluate the effect of 3D positions and image depth; (4) {\bf Base+VFE+DT}: a variant model of MEP3P w/o MLLM employs VFE and DT to evaluate the effect of image depth and depth-aware attention; (5)
{\bf Base+VFE+DT+POMRC}: the proposed MEP3P w/o MLLM model incorporates VFE, DT and POMRC; (6) {\bf MLLM}: a model only adopts BLIP-2 to process image and text after referring expressions replaced, and feeds the pooled output of BLIP-2 to the classification network for reasoning, which is to evaluate the ability of MLLM to do VCR; (7) {\bf Base+MER}: the Base model further uses MER to perform VCR, which is to evaluate the effect of MER; and (8) {\bf Base+VFE+DT+POMRC+MER}: the proposed MEP3P incorporates VFE, DT, POMRC and MER.

\begin{table}[htbp]
\renewcommand\arraystretch{1.1}
\centering
\caption{Ablation study on the VCR validation set.}
\label{tab2}
\begin{lrbox}{\tableboxTabOV}
\begin{tabular}{c|ccc}
\toprule[1pt]
Methods  & $Q\rightarrow A$ & $QA\rightarrow R$ & $Q\rightarrow AR$ \\
\hline
Base & 70.2 & 71.6 & 50.5 \\
Base+VFE2D & 70.6 & 72.1 & 51.1  \\
Base+VFE & 71.2 & 72.6 & 51.9  \\
Base+VFE+DT & 71.8 & 73.3 & 52.9  \\
Base+VFE+DT+POMRC & 72.2 & 73.8 & 53.5  \\\hline
MLLM & 62.8 & 57.5 & 37.0  \\
Base+MER & 75.4 & 74.7 & 56.6  \\
Base+VFE+DT+POMRC+MER & {\bf77.6} & {\bf77.0} & {\bf60.0}  \\
\toprule[1pt]
\end{tabular}
\end{lrbox}
\scalebox{1.0}{\usebox{\tableboxTabOV}}
\end{table}

\begin{figure*}[htbp]
   \centering
   \includegraphics [width=0.95\linewidth] {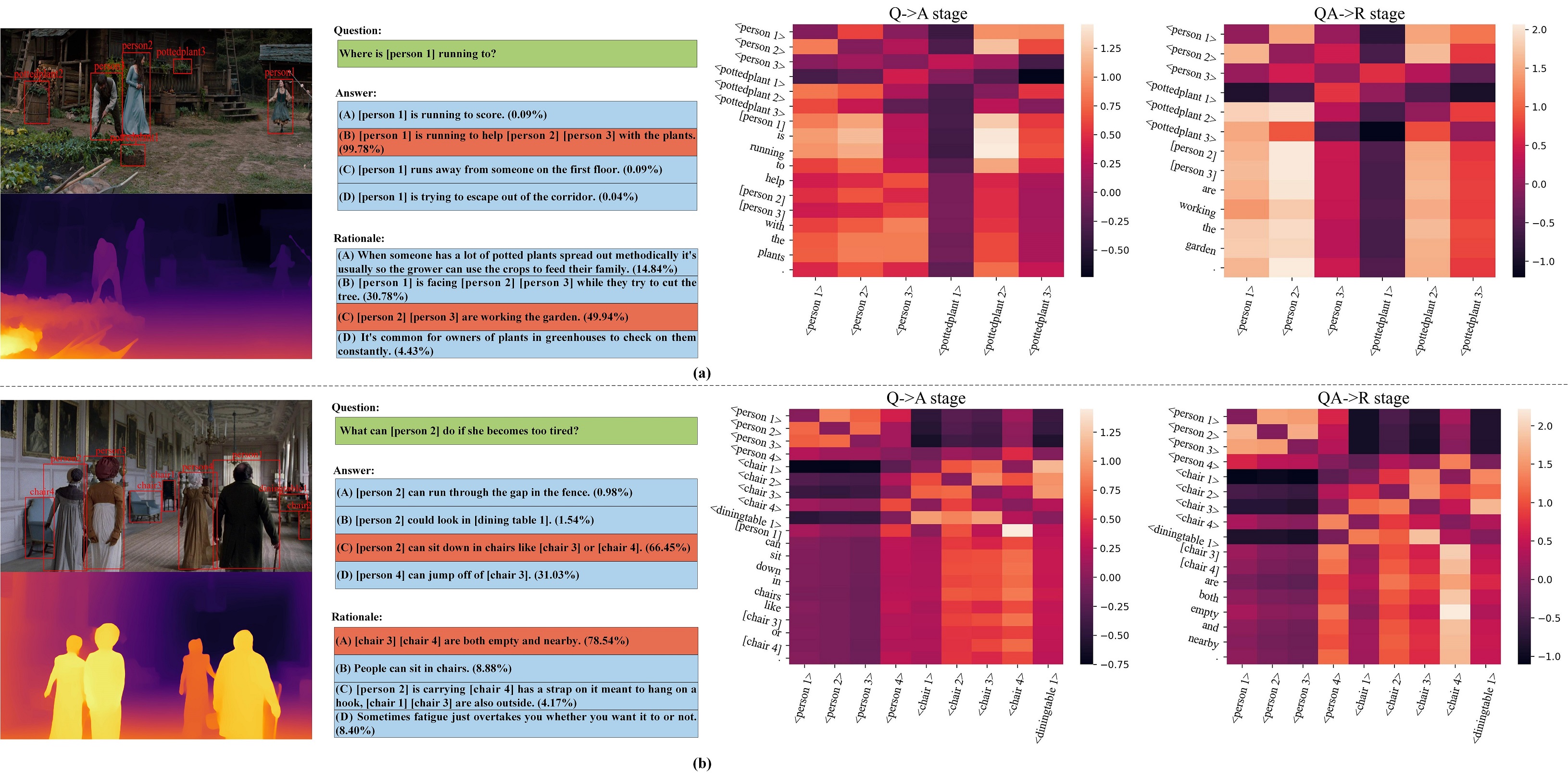}
   \caption{Instances obtained by the proposed MEP3P w/o MLLM. The percentages in brackets are the probabilities predicted by MEP3P w/o MLLM, and the choices filled in brown are ground-truths. To be distinguishable, $<\cdot>$ means visual object and $[\cdot]$ means linguistic entity. The heatmaps for the ground-truth choices on the right indicate the final adjustment matrice for answer-image contextualization.
   } \label{fig4}
\end{figure*}

\begin{figure*}[htbp]
   \centering
   \includegraphics [width=0.9\linewidth] {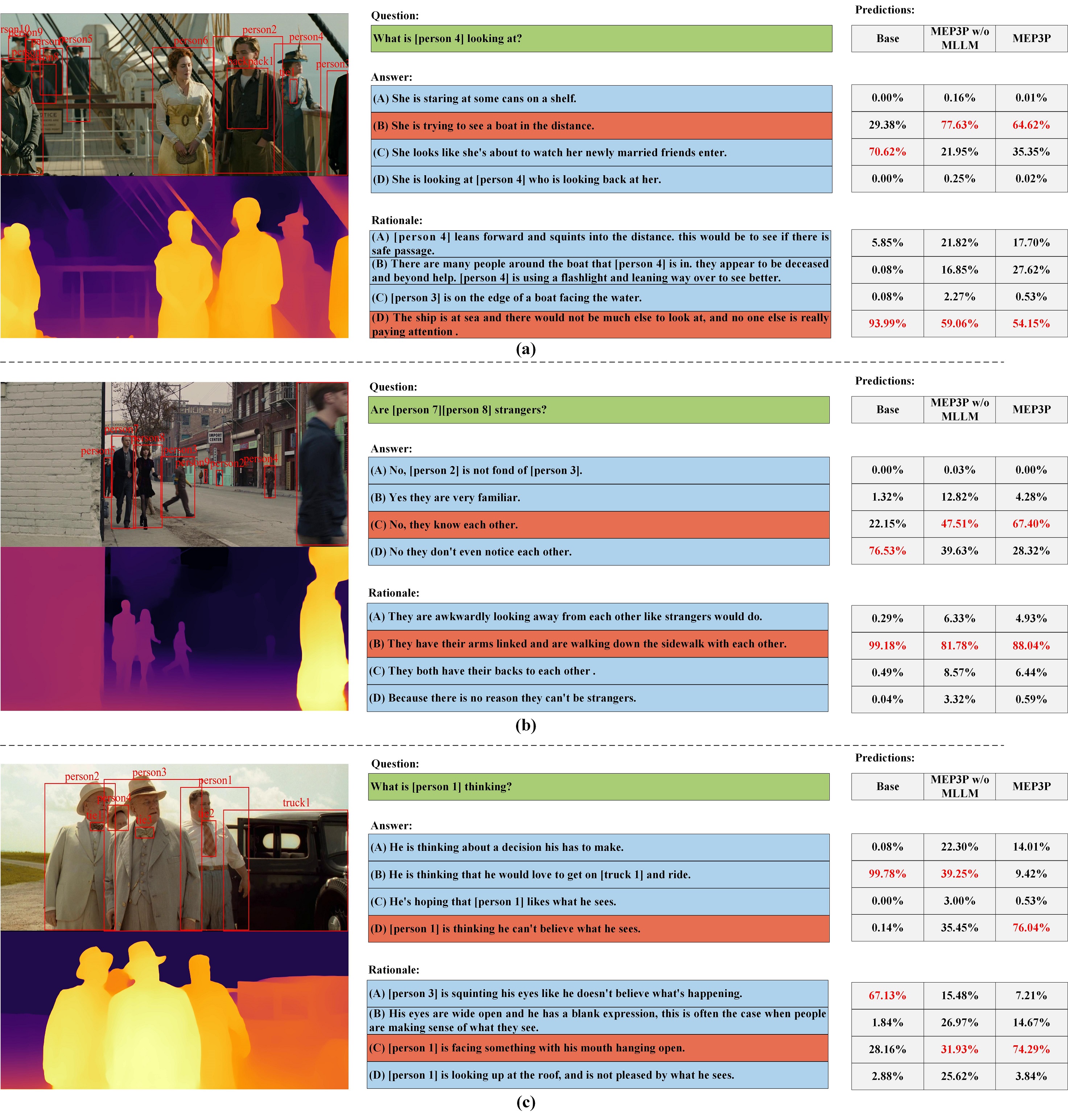}
   \caption{Illustration of prediction results obtained by the Base model, MEP3P w/o MLLM and MEP3P. The choices filled in brown are ground-truths and the predictions achieved by different models are shown on the right.
   } \label{fig5}
\end{figure*}

The results of ablation study for the three VCR subtasks are shown in Table~\ref{tab2}. As can be observed, Base+VFE2D obtains the improvements of 0.4\% for $Q\rightarrow A$, 0.5\% for $QA\rightarrow R$, and 0.6\% for $Q \rightarrow AR$, respectively, indicating the importance of object positions for image understanding. With depth information considered, Base+VFE further gains improvements since pseudo 3D positions are more exact. Compared with Base+VFE, Base+VFE+DT adopting depth difference to guide attention achieves better performances of 0.6\%, 0.7\% and 1.0\% improvements for the three subtasks, indicating that depth difference makes the model to pay more attention to the objects with the related depth level. The proposed MEP3P w/o MLLM utilizes POMRC to optimize parameters and obtains better performances. As for MLLM, the performance obtained is poor since the model cannot distinguish visual objects referred to. Particularly, most VCR frameworks perform better in $QA\rightarrow R$ compared with $Q\rightarrow A$ while MLLM performs worse in $QA\rightarrow R$. This is because more information is provided in the $QA\rightarrow R$ task to do reasoning for VCR models.  However, when calculating aligned cross-modal features, MLLM combines identical components~(\emph{i.e.}, image, question and answer) with different rationales as the input to judge the four choices in $QA\rightarrow R$. Similarly, MLLM combines identical components~(\emph{i.e.}, image and question) with different answers in $Q\rightarrow A$. The representation for $QA\rightarrow R$ contains more identical components, making it more difficult to be distinguished compared to $Q\rightarrow A$. With the assistance of the Base model which owns referring capability, Base+MER employs aligned cross-modal features and gains improvements by a large margin. Besides MER, the proposed MEP3P additionally utilizes the VFE, DT and POMRC modules and obtains the best performance.

\subsection{Visualization Analysis}
\label{sect:exp:vis}

Figure~\ref{fig4} shows two typical instances obtained by the proposed MEP3P w/o MLLM. As can be seen in Fig.~\ref{fig4}(a), the correlation between irrelevant $<person~1>$ and $<pottedplant~1>$ is weakened in the adjustment matrix guided by depth difference. The answer and rational both capture the depth level of $<person~1>$, $<person~2>$, $<person~3>$ and $ <pottedplant~2>$. In Fig.~\ref{fig4}(b), the answer and rational both pay more attention to the chairs with similar depths~(\emph{i.e.}, $<chair~2>$, $<chair~3>$ and $<chair~4>$ except $<chair~1>$) based on the understanding of the visual scene. This is because the proposed MEP3P w/o MLLM can properly focus on key cues. Although the models for $Q\rightarrow A$ and $QA\rightarrow R$ are trained separately, the elements in the adjustment matrice from object to word appear similar, which demonstrates the stability of the proposed depth-aware attention.

Three typical prediction results obtained by the Base model, MEP3P w/o MLLM and MEP3P are illustrated in Fig.~\ref{fig5}. In Fig.~\ref{fig5}(a), the Base model without position features incorrectly associates $ [person~4]$ with the visual objects on the boat. The proposed MEP3P w/o MLLM and MEP3P both know that $[person~4]$ is on the edge of the image and far from most visual objects. Therefore, MEP3P w/o MLLM and MEP3P with position features achieve the correct prediction. In Fig.~\ref{fig5}(b), the Base model doesn't know that $[person~7]$ and $[person~8]$ are neighbouring and fails to understand the semantic, while MEP3P w/o MLLM and MEP3P correctly associate $[person~7]$ with $[person~8]$. In Fig.~\ref{fig5}(c), both the Base model and MEP3P w/o MLLM insufficiently understand the sematic, and MEP3P with the assistance of MLLM can precisely judge the matching of cross-modal semantic.

\subsection{Extension on Proposed Module}
\label{sect:exp:etb}

The proposed VFE, depth-aware attention~(DA), MER and POMRC modules are the key components in MEP3P. These components can be conveniently extended to other VCR methods, with the results shown in Table~\ref{tab3}. Note that all the results reported are reproduced from the official implementations~\cite{Yu2019Heterogeneous, Lin2019TAB, Zhang2021Multi}. As can be observed, using VFE2D can help HGL~\cite{Yu2019Heterogeneous} and TAB-VCR~\cite{Lin2019TAB} to gain around 0.4\%, 0.5\% and 0.7\% improvements for the three subtasks. The performances can be further boosted by using VFE instead of VFE2D. Since MCC~\cite{Zhang2021Multi} has adopted 2D positions to distinguish samples in contrastive learning, the performances achieved by MCC+VFE are reported. DA can be further utilized to associate visual objects with similar depth values, and around 0.5\%, 0.5\% and 0.7\% improvements are achieved for the three subtasks. With the assistance of MER, all the three baselines gain improvements by a large margin. Moreover, all the three baselines achieve improvements with POMRC employed. Finally, these baselines are extended with the proposed modules jointly used, and an average improvement of around 9\% is achieved for the $Q \rightarrow AR$ task. This indicates that the proposed modules can be applicable to other VCR methods.

\begin{table*}[htbp]
\renewcommand\arraystretch{1.1}
\centering
\caption{Performances of extending proposed modules to three VCR baselines.}
\label{tab3}
\begin{lrbox}{\tableboxTabOV}
\begin{tabular}{c|ccccc|ccc}
\toprule[1pt]
Models  &VFE2D &VFE &DA & MER &POMRC & $Q\rightarrow A$ & $QA\rightarrow R$ & $Q\rightarrow AR$ \\
\hline
HGL~\cite{Yu2019Heterogeneous}  &- &- &- &- &- & 69.3 & 70.6 & 49.0 \\
HGL+VFE2D &\ding{51} & &- &- &- & 69.7 & 71.1 & 49.7  \\
HGL+VFE &- &\ding{51} &- &- &- & 70.2 & 71.7 & 50.6  \\
HGL+VFE+DA &- &\ding{51} &\ding{51} &- &- & 70.7 & 72.2 & 51.3  \\
HGL+MER &- &- &- &\ding{51} &-  & 74.3 & 73.1 & 54.2  \\
HGL+POMRC &- &- &- &- &\ding{51}  & 69.9 & 71.3 & 49.9  \\
HGL+VFE+DA+POMRC+MER &- &\ding{51} &\ding{51} &\ding{51} &\ding{51}  & 76.3 & 75.2 & 57.4  \\
\hline
TAB-VCR~\cite{Lin2019TAB} &- &- &- &- &- & 69.7 & 71.9 & 50.2 \\
TAB-VCR+VFE2D &\ding{51} & &- &- &- & 70.1 & 72.3 & 50.8  \\
TAB-VCR+VFE &- &\ding{51} &- &- &- & 70.6 & 72.8 & 51.5  \\
TAB-VCR+VFE+DA &- &\ding{51} &\ding{51} &- &- & 71.1 & 73.4 & 52.3  \\
TAB-VCR+MER &- &- &- &\ding{51} &-  & 74.9 & 74.5 & 55.9  \\
TAB-VCR+POMRC &- &- &- &- &\ding{51}  & 70.3 & 72.5 & 51.0  \\
TAB-VCR+VFE+DA+POMRC+MER &- &\ding{51} &\ding{51} &\ding{51} &\ding{51}  & 77.0 & 76.7 & 59.1  \\
\hline
MCC~\cite{Zhang2021Multi} &- &- &- &- &- & 70.9 & 72.1 & 51.2 \\
MCC+VFE &- &\ding{51} &- &- &- & 71.4 & 72.7 & 52.0  \\
MCC+VFE+DA &- &\ding{51} &\ding{51} &- &- & 71.9 & 73.2 & 52.7  \\
MCC+MER &- &- &- &\ding{51} &-  & 76.0 & 75.2 & 57.3  \\
MCC+POMRC &- &- &- &- &\ding{51}  & 71.3 & 72.5 & 51.8  \\
MCC+VFE+DA+POMRC+MER &- &\ding{51} &\ding{51} &\ding{51} &\ding{51}  & 78.1 & 77.5 & 60.7  \\
\toprule[1pt]
\end{tabular}
\end{lrbox}
\scalebox{1.0}{\usebox{\tableboxTabOV}}
\end{table*}

\subsection{Parameter Analysis}
\label{sect:exp:pa}

\begin{figure*}[htbp]
   \centering
   \includegraphics [width=1.0\linewidth] {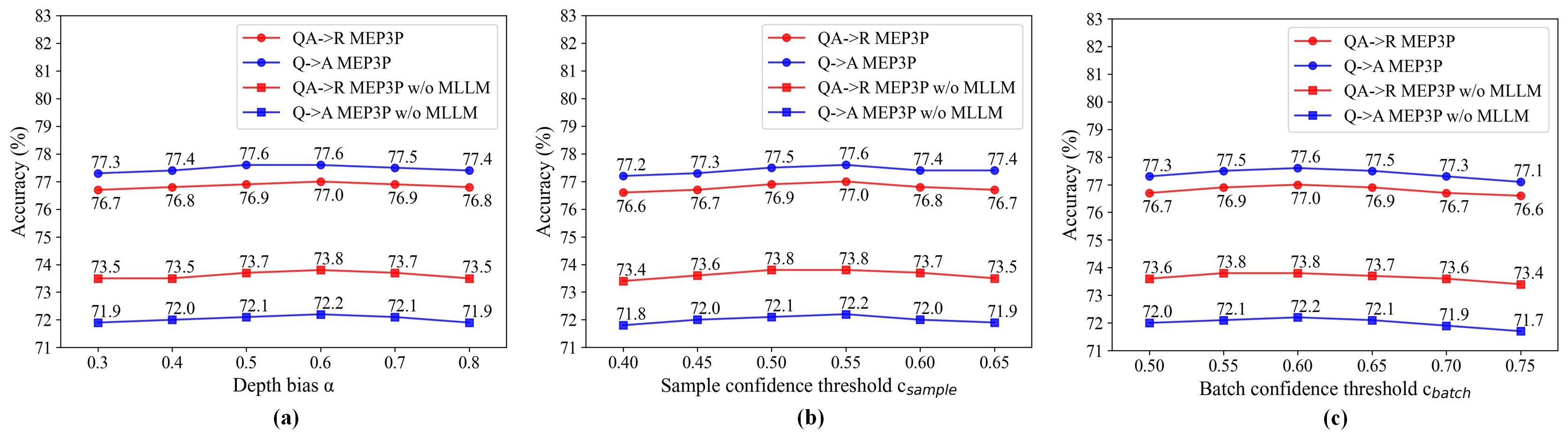}
   \caption{Accuracy trends achieved by the proposed frameworks with (a)~different $\alpha$ values, (b)~different $c_{sample}$ values and (c)~different $c_{batch}$ values.
   } \label{fig6}
\end{figure*}

\subsubsection{Depth Bias}

To investigate the effect of the depth bias $\alpha$, the performances achieved by the proposed frameworks with different $\alpha$ values are shown in Fig.~\ref{fig6}(a). As illustrated, the reasoning accuracy declines when the attention to most objects is strengthened by too small $\alpha$. On the other hand, the attention to a few valuable objects is weakened when $\alpha$ is too big, which also leads to worse performance. Therefore, $\alpha$ is set as 0.6 to make the model emphasize valuable objects appropriately.

\subsubsection{Multi-level Confidence Threshold}

The multi-level confidence thresholds $c_{sample}$ and $c_{batch}$ are used to select high-quality optimization sources in sample-level and batch-level, respectively. As is illustrated in Fig.~\ref{fig6}(b), POMRC degenerates with most correct samples selected when $c_{sample}$ is too small. On the other hand, if $c_{sample}$ is too big, the sample selection principle will be too strict with too few samples selected. As can be observed in Fig.~\ref{fig6}(c), $c_{batch}$ also needs to be set suitably to select moderate high-quality batches. As a result, $c_{sample}$ and $c_{batch}$ are set to 0.55 and 0.6, respectively, achieving the best performance.

\section{Conclusion}
\label{sect:cln}

In this work, a novel MEP3P method is developed to perform the VCR task, which consists of two essential designs: P3PV and MER. Specifically, we firstly demonstrate the relation between objects is relevant to object depth in images. P3PV is designed to represent the relation of visual objects in a pseudo-3D manner with image depth introduced and associate objects and answer words with visual scenes guided by depth. In the reasoning stage, MER is proposed to cooperate with P3PV by providing well-aligned cross-modal features and referring capability. To better optimize model parameters, POMRC is devised to consider the batch quality and integrate multi-step optimization results. The experiments conducted on the benchmark VCR dataset demonstrate the effectiveness of the proposed method compared with several state-of-the-art approaches.

In the future work, the proposed depth-aware Transformer can be applied to 3D tasks with precise depth information such as scene understanding in autonomous driving. In addition, more effective techniques will be studied to integrate fine-grained knowledge of specific tasks into MLLMs. Furthermore, a unified framework is worth investigated to perform various vision-and-language tasks via the proposed MLLM enhanced pseudo 3D perception method.

\bibliographystyle{IEEEtran}
\bibliography{ref}

%




\end{document}